\documentclass[10pt,twocolumn,letterpaper]{article}

\usepackage{wacv}
\usepackage{times}
\usepackage{epsfig}
\usepackage{graphicx}
\usepackage{amsmath}
\usepackage{amssymb}
\usepackage{booktabs}
\usepackage{multirow}
\usepackage{collcell}
\usepackage{paralist}
\usepackage[accsupp]{axessibility}  

\def\wacvPaperID{486} 

\wacvfinalcopy 

\ifwacvfinal
\def\assignedStartPage{1}
\fi

\ifwacvfinal
\usepackage[pagebackref=true,breaklinks=true,colorlinks,bookmarks=false,citecolor=blue,linkcolor=blue]{hyperref}
\else
\usepackage[pagebackref=true,breaklinks=true,letterpaper=true,colorlinks,bookmarks=false,citecolor=blue,linkcolor=blue]{hyperref}
\fi

\ifwacvfinal
\else
\fi

\def\eg{\emph{e.g.,~}}               
\def\ie{\emph{i.e.,~}}               

\begin{document}

\title{Video Salient Object Detection via Contrastive Features and Attention Modules}

\author{Yi-Wen Chen$^1$\thanks{Work done during an internship at ByteDance AI Lab.} \hspace{5mm}  Xiaojie Jin$^2$\thanks{Corresponding author.} \hspace{5mm}  Xiaohui Shen$^2$ \hspace{5mm} Ming-Hsuan Yang$^{1}$ \\
$^1$University of California at Merced \hspace{5mm}  $^2$ByteDance AI Lab}
\maketitle

\begin{abstract}
Video salient object detection aims to find the most visually distinctive objects in a video.
To explore the temporal dependencies, existing methods usually resort to recurrent neural networks or optical flow. However, these approaches require high computational cost, and tend to accumulate inaccuracies over time.
In this paper, we propose a network with attention modules to learn contrastive features for video salient object detection without the high computational temporal modeling techniques. We develop a non-local self-attention scheme to capture the global information in the video frame. A co-attention formulation is utilized to combine the low-level and high-level features.
We further apply the contrastive learning to improve the feature representations, where foreground region pairs from the same video are pulled together, and foreground-background region pairs are pushed away in the latent space.
%
The intra-frame contrastive loss helps separate the foreground and background features, and the inter-frame contrastive loss improves the temporal consistency.
We conduct extensive experiments on several benchmark datasets for video salient object detection and unsupervised video object segmentation, and show that the proposed method requires less computation, and performs favorably against the state-of-the-art approaches.
\end{abstract}


\vspace{-4mm}
\section{Introduction}
\vspace{-1mm}
Video saliency detection is a task to find the most visually attractive regions in each frame of a video. It is a fundamental technique in computer vision and can be used for many high-level applications, such as video object segmentation, visual object tracking and video editing.
Research work on saliency detection can be roughly classified into two groups, \ie eye fixation prediction \cite{Wang_CVPR_2018_2,Min_ICCV_2019} and salient object detection \cite{Wang_TIP_2018,Song_ECCV_2018,Fan_CVPR_2019,Li_ICCV_2019}. The former is to predict the locations in a scene where a human observer may fixate, while the latter aims to uniformly highlight the most salient object regions. In this paper, we focus on the salient object detection task in videos.
The challenging part of this task is to distinguish the primary object from the background, as well as keep the temporal correspondence.
Early methods for video saliency detection are built upon hand-crafted features, such as color and texture. These models are limited by the representation ability of low-level features. With the success of deep CNN models in many computer vision tasks, recent approaches improve the performance of video saliency detection by learning fully convolutional networks (FCNs).
In order to model the temporal information, optical flow is often utilized to provide the motion cues. However, the performance of these methods highly relies on the accuracy of flow computation, and is vulnerable to conditions where the foreground objects are nearly static.
Other methods \cite{Song_ECCV_2018,Fan_CVPR_2019} adopt models based on recurrent neural networks such as ConvLSTM to aggregate the spatiotemporal information. While these approaches achieve promising performance, the computational cost is significantly high for modeling temporal information.

\begin{figure}[t]
	\centering
	\includegraphics[width=.8\linewidth]{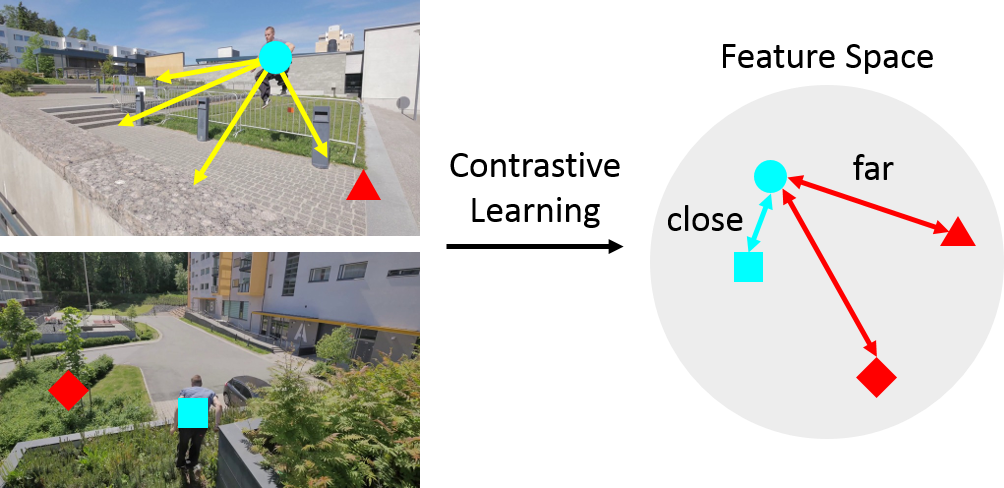}
	\caption{\textbf{Overview of the proposed algorithm.} We use the non-local self-attention (illustrated as the yellow lines) to capture the long-distance correspondence, where the features of a location are computed upon all the other locations. To learn contrastive features from the videos, for each foreground region anchor (circle), we select foreground regions from the same video as positive samples (square), and background regions as negative samples (triangle and diamond).
	}
	\label{fig:overview}
	\vspace{-6mm}
\end{figure}

In this paper, we propose a framework with attention modules to learn contrastive features for video salient object detection. The main ideas are illustrated in Figure \ref{fig:overview}.
To capture the long-range dependencies in video frames, we develop a non-local self-attention module that directly computes the interactions between any two positions similar to the non-local operation in \cite{Wang_CVPR_2018}.
%
Since the non-local operation in \cite{Wang_CVPR_2018} is computationally expensive, we develop an efficient scheme where the feature transformations are shared, and the $1 \times 1$ convolution is replaced with the $3 \times 3$ depthwise convolution.
We also incorporate the dynamic convolution into our model, where filters could learn to adapt to the input.
To better utilize the features from different levels of the deep network, we design a cross-level co-attention mechanism to explore the correspondence between low-level and high-level features. 
This approach helps the model focus on the most distinctive regions in the scene while maintaining the high-resolution outputs, which is more effective than previous feature fusion mechanisms such as simply concatenating features from different levels.

To help the model learn discriminative features between foreground and background regions, we integrate the contrastive learning technique into our model. Intuitively, the foreground regions in the same video should share similar features, while foreground and background are supposed to have different features. 
Based on this criterion, for each foreground region anchor, we select the foreground regions in the same video as positive samples, and background regions as negative samples. The objective of the contrastive loss is to minimize the distance between the anchor and positive samples in the embedding space, while maximizing the distance between the anchor and negative samples.
To provide stronger supervision, we introduce two hard sample mining strategies to find meaningful positive and negative samples.
By sampling from the whole video, the intra-frame contrastive loss helps the model learn distinctive features of foreground and background, and the inter-frame contrastive loss improves the temporal correspondence of the network.
%
While contrastive learning is widely used in self-supervised representation learning, it has not been exploited in video salient object detection.
Different from recent contrastive learning approaches \cite{info_nce,moco,simclr} that consider one positive sample augmented from the same image, and multiple negative samples from different images, in our method, we select multiple positive and negative samples, where the samples are generated from foreground and background regions in the same video.

To evaluate the proposed method, we conduct extensive experiments on several video saliency benchmark datasets. We also apply the proposed framework on the unsupervised video object segmentation task as an example of applications.
Experimental results show that our method performs favorably against the state-of-the-art approaches.
The main contributions of this work are summarized as follows:
\begin{compactitem}
\item We propose an efficient framework for video salient object detection that achieves state-of-the-art performance with \emph{higher accuracy}, \emph{smaller model size} and \emph{lower runtime}. To date, there is no other method that can accomplish all three goals.
\item We design a non-local self-attention operation that captures global information, and a cross-level co-attention formulation to explore the correspondence between low-level and high-level features.
\item We integrate the contrastive learning technique and hard sample mining strategy into our network to learn contrastive features between foreground and background regions, and improve the temporal consistency across video frames.
\end{compactitem}







\vspace{-1mm}
\section{Related Work}
\vspace{-2mm}
{\flushleft {\bf Video Salient Object Detection.}}
Video salient object detection aims to find the most distinctive object regions in each video frame. Conventional methods usually extract hand-crafted features in spatial and temporal dimensions separately and then integrate them together.
They distinguish the salient objects from background by classic heuristics such as color contrast~\cite{Achanta_CVPR_2009_2}, background prior~\cite{Yang_CVPR_2013} and center prior~\cite{Jiang_CVPR_2013}. Optical flow is often utilized for exploring temporal information. The performance of these approaches is limited by the representation ability of the low-level features. Furthermore, the computational cost is high for optical flow estimation.

As the deep CNN models achieved success in many computer vision tasks, recent approaches adopt CNN-based models for video salient object detection and improve the results.
The FCNS~\cite{Wang_TIP_2018} method employs a static FCN for single frame saliency prediction, and another dynamic FCN that takes the initial saliency map and the consecutive frame pairs as input to explore temporal information.
Other approaches model the spatiotemporal information by 3D convolutional filters~\cite{Le_BMVC_2017}, ConvLSTM~\cite{Li_CVPR_2018_3,Song_ECCV_2018,Fan_CVPR_2019} or jointly considering appearance and optical flow~\cite{Li_CVPR_2018_3,Li_ICCV_2019} in the framework.
While these CNN-based networks generally achieve state-of-the-art results, they usually require high computational cost for modeling temporal information, and the temporal dependencies are often limited in short time intervals, which may accumulate errors over time.

\begin{figure*}[tp]
	\centering
	\includegraphics[width=.75\linewidth]{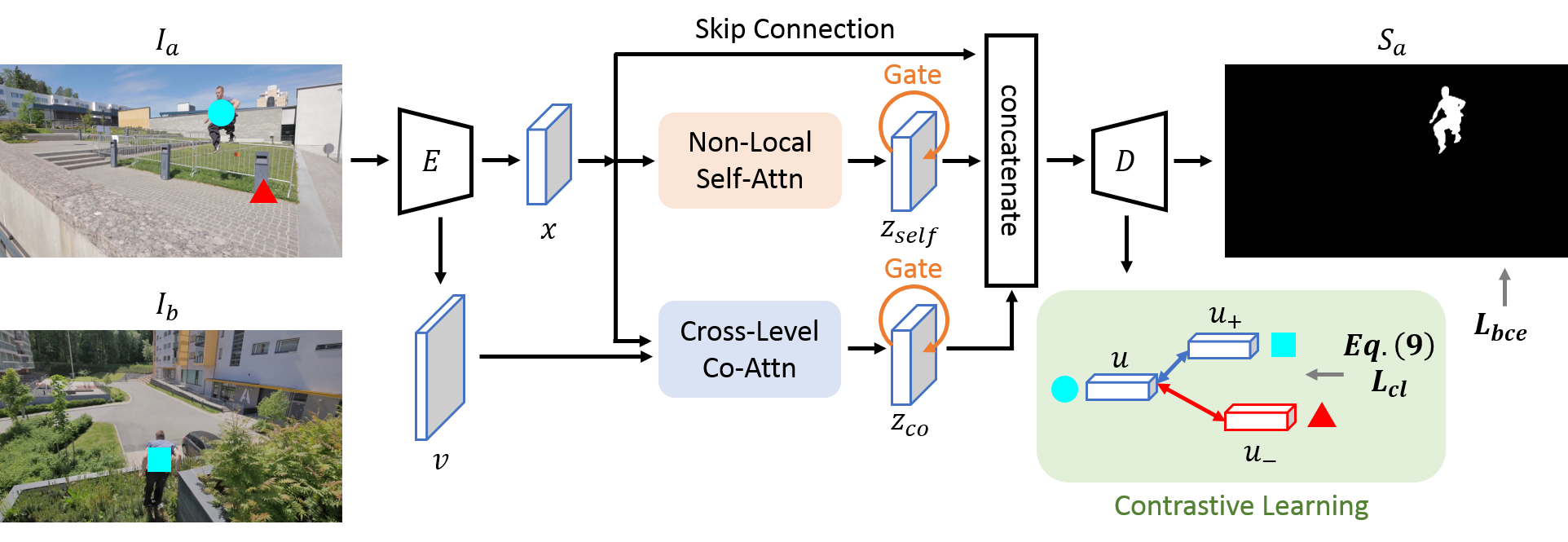}
	\vspace{-2mm}
	\caption{\textbf{Pipeline of the proposed framework.} The feature of the input image $I_a$ is first extracted by an encoder $E$. Then we apply a non-local self-attention module on the feature $\mathbf{x}$, and generate the self-attended feature $\mathbf{z}_{self}$. A cross-level co-attention module takes the low-level feature $\mathbf{v}$ and the high-level feature $\mathbf{x}$ as input, and produces the co-attended feature $\mathbf{z}_{co}$. The attentive features $\mathbf{z}_{self}$ and $\mathbf{z}_{co}$ are weighted by a gating function, respectively.
	Finally, the original feature $\mathbf{x}$, self-attended feature $\tilde{\mathbf{z}}_{self}$ and co-attended feature $\tilde{\mathbf{z}}_{co}$ are concatenated and fed into the detection head $D$ to generate the output saliency map $S_a$. To learn better feature representations, we apply the contrastive loss $L_{cl}$ on the features before the final output layer. For an anchor foreground region with feature $u$, we select foreground regions with feature $u_+$ as positive samples, and background regions with feature $u_-$ as negative samples. The samples are chosen from frames in the same video to improve the intra-frame discriminability as well as inter-frame consistency.}
	\label{fig:framework}
	\vspace{-4mm}
\end{figure*}

\vspace{-2mm}
{\flushleft {\bf Unsupervised Video Object Segmentation.}}
The task of video object segmentation is typically tackled in either the \emph{semi-supervised} or \emph{unsupervised} setting. In the semi-supervised setting, the ground truth mask of the first frame is given during the testing stage, while no supervision is given during testing in the unsupervised setting.
Unsupervised video object segmentation is similar to video salient object detection. The difference is that the output is a binary map in the segmentation task, but a continuous-valued map in the saliency task.
Conventional methods for unsupervised video object segmentation are based on hand-crafted features, and utilize techniques such as object proposal ranking~\cite{Perazzi_ICCV_2015,Koh_CVPR_2017}, trajectory clustering~\cite{Ochs_ICCV_2011,Keuper_ICCV_2015}, motion analysis~\cite{Papazoglou_ICCV_2013,Tokmakov_CVPR_2017}, and saliency information~\cite{Wang_CVPR_2015,Wang_CVPR_2019}.

Recent approaches \cite{Chen_ACCV_2018,Chen_IJCV_2020,Zhou_AAAI_2020} focus on learning deep CNN models and employ motion cues to keep the temporal consistency. The FSEG~\cite{Jain_CVPR_2017} method adopts a two-stream network to jointly consider the appearance and motion information. The SegFlow~\cite{Cheng_ICCV_2017} mechanism shows that joint learning the video object segmentation and optical flow can improve the performance of both tasks. However, these approaches are still limited by drawbacks of optical flow that requires high computational cost and only considers short-term temporal dependencies.
To tackle this problem, some other approaches capture global information of the whole video without computing optical flow. The COSNet~\cite{Lu_CVPR_2019} method uses a co-attention mechanism to model the correlation of randomly selected frames from the input video.
The AGNN~\cite{Wang_ICCV_2019} scheme builds a fully connected graph to represent video frames as nodes, and the pair-wise relationships between frames as edges.
In the AnDiff~\cite{Yang_ICCV_2019_3} model, the first frame is considered as the anchor, whose features are propagated to the current frame via an aggregation technique.
The DFNet~\cite{Zhen_ECCV_2020} approach learns discriminative features under CRF formulation with several frames sampled from the input video.
While we use a similar co-attention scheme as COSNet, our motivation and usage are different from theirs.
COSNet performs co-attention between frames during both training and testing to explore the temporal relevance among frames. It requires five reference frames in the testing stage, which largely increases runtime.
In contrast, we apply co-attention between different levels of features to incorporate the high-resolution details into the semantic features.
Our temporal information is explored in the contrastive learning technique, which is more efficient.

\vspace{-1mm}
\section{Proposed Framework}
\vspace{-1mm}
In this work, we focus on detecting the most salient object regions in videos. Given an input video $I = \{I_1, ..., I_n\}$ with $n$ frames, we aim to generate a sequence of saliency maps $S = \{S_1, ..., S_n\}$ with values in the range $[0, 1]$.
To this end, we propose an end-to-end trainable network that contains a feature encoder, two attention modules and a detection head. The first attention technique is the non-local self-attention that captures long-range dependencies in the features. The second attention mechanism is the cross-level co-attention which finds the correspondence between the low-level and high-level features to generate saliency maps with higher quality.

Figure~\ref{fig:framework} shows the pipeline of the proposed framework.
We first extract the feature representations of the input frame using the feature encoder $E$. Then two attention modules aggregate the features from different perspectives. The attentive features $\mathbf{z}_{self}$ and $\mathbf{z}_{co}$ are weighted by a gating function. Finally, the detection head $D$ takes the concatenation of the original feature $\mathbf{x}$, self-attended feature $\tilde{\mathbf{z}}_{self}$ and co-attended feature $\tilde{\mathbf{z}}_{co}$ to generate the saliency map.
To learn better feature representations for distinguishing foreground from background, we integrate the contrastive learning technique into our model. The objective is to make features of foreground regions in the same video close to each other, and far away from the background regions.
By selecting the positive and negative samples from the whole video, the model is able to learn contrastive features to separate foreground from background, as well as improve the temporal correspondence.


\begin{figure}[t]
	\centering
	\includegraphics[width=.9\linewidth]{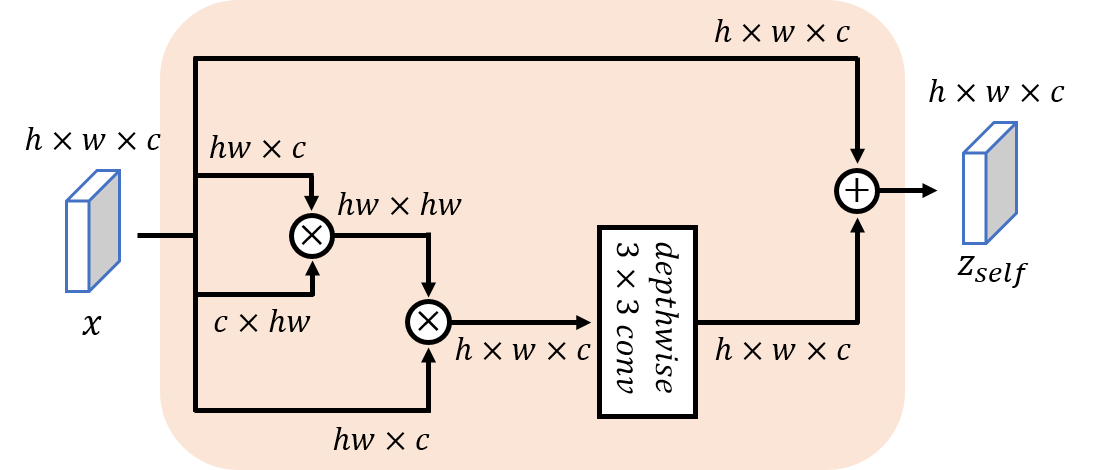}
	\caption{\textbf{Non-local self-attention module.} $\otimes$ represents matrix multiplication, and $\oplus$ denotes element-wise summation. We show the input shape for each operation.}
	\label{fig:self_attn}
	\vspace{-4mm}
\end{figure}

\subsection{Non-Local Self-Attention}
Although CNN-based models have shown great success on many computer vision tasks, one limitation of the convolutional operation is that it processes only one local neighborhood at a time. To consider long-distance dependencies in images, convolutional operations are stacked to model large receptive fields.
However, repeating local operations is computationally inefficient, and will cause optimization difficulties. In order to capture long-range dependencies in the video frames more efficiently, we develop the non-local self-attention technique similar to the non-local operation in \cite{Wang_CVPR_2018}.
Specifically, the non-local self-attention aims to find the correlations of all positions by directly computing the relationships between any two positions in the input frame.
Given the feature representation $\mathbf{x}$ extracted from the input frame, we perform the non-local operation:
\begin{equation}
\mathbf{y}_i = \frac{1}{\mathcal{C}(\mathbf{x})} \sum\limits_{\forall j} f(\mathbf{x}_i, \mathbf{x}_j) g(\mathbf{x}_j),
\label{eq:non_local}
\end{equation}
where $i$ is the index of an output position, $j$ enumerates all possible positions in $\mathbf{x}$, $f(\mathbf{x}_i, \mathbf{x}_j)$ represents the affinity matrix between $\mathbf{x}_i$ and its context features $\mathbf{x}_j$, $g(\mathbf{x}_j)$ computes a representation of the feature at position $j$, and $\mathcal{C}(\mathbf{x})$ is the normalization factor.
The non-local operation in \eqref{eq:non_local} is then wrapped into a non-local block with a residual connection:
\begin{equation}
\mathbf{z}_i = W_z \mathbf{y}_i + \mathbf{x}_i,
\label{eq:non_local_block}
\end{equation}
where $W_z$ is a learnable matrix for feature transformation.
There are multiple choices for the functions $f(\cdot, \cdot)$ and $g(\cdot)$. A simple version is to apply dot product for $f(\cdot, \cdot)$, and a linear embedding (\eg $1 \times 1$ convolution) for $g(\cdot)$. Then \eqref{eq:non_local} can be written as:
\begin{equation}
\mathbf{y} = \frac{1}{\mathcal{C}(\mathbf{x})} \theta(\mathbf{x}) \phi(\mathbf{x})^\top g(\mathbf{x}),
\label{eq:non_local_dot}
\end{equation}
where $\mathbf{x}$ is of shape $(h, w, c)$, with $h$, $w$ and $c$ denoting the height, width and number of channels, respectively. $\theta(\cdot)$, $\phi(\cdot)$ and $g(\cdot)$ are $1 \times 1$ convolutional layers with $c$ channels, and $\top$ represents the transpose operation. The outputs of the convolutional layers are reshaped to $(h \times w, c)$ before matrix multiplications.
%
%
The normalization factor $\mathcal{C}(\mathbf{x})$ is set as $N$, which is the number of positions in $\mathbf{x}$.
Here we employ a lightweight version of non-local networks as in \cite{Li_CVPR_2020}, where the feature transformations $\theta(\cdot)$, $\phi(\cdot)$ and $g(\cdot)$ in \eqref{eq:non_local_dot} are shared and integrated into the feature extractor that produces $\mathbf{x}$. Therefore, \eqref{eq:non_local_dot} is simplified as:
\begin{equation}
\mathbf{y} = \frac{1}{N} \mathbf{x} \mathbf{x}^\top \mathbf{x}.
\label{eq:non_local_light}
\end{equation}
In the recent work \cite{Wang_CVPR_2018}, the feature transformation $W_z$ in \eqref{eq:non_local_block} is instantiated as a $1 \times 1$ convolutional layer, which is a heavy computation. Therefore, we replace the $1 \times 1$ convolution with $3 \times 3$ depthwise convolution for higher efficiency. \eqref{eq:non_local_block} is then revised as:
\begin{equation}
\mathbf{z}_{self} = DepthwiseConv(\mathbf{y}, W_d) + \mathbf{x},
\label{eq:non_local_self}
\end{equation}
where $\mathbf{z}_{self}$ is the self-attended feature, and $W_d$ is the depthwise convolution kernel.
We implement the depthwise convolution as a dynamic convolution operation~\cite{Brabandere_NIPS_2016}. Instead of applying the same set of weights to all input images, dynamic convolutional filters are generated depending on the features of the input image, which equips the model with better adaptability to the input.
Specifically, we adopt a single convolutional layer that takes the feature $\mathbf{x}$ as input to generate the dynamic convolutional filters $W_d$.
The non-local self-attention operation is illustrated in Figure~\ref{fig:self_attn}.

%
%
%
%
%

\subsection{Cross-Level Co-Attention}
In deep CNN models, layers near the input contain low-level features, while those near the output represent high-level features. In order to find the salient objects using high-level semantic features as well as keep the details from low-level features, we employ a cross-level co-attention module to explore the relationships between features from two different levels.
With the features generated from the feature extractor, we select one low-level feature $\mathbf{v}$ and one high-level feature $\mathbf{x}$. To mine the correlations between the two features with different sizes, we first adopt a convolutional layer to make the size of $\mathbf{v}$ same as $\mathbf{x}$.
Then we compute the affinity matrix from the two features:
\begin{equation}
A = \mathbf{v} W \mathbf{x}^\top,
\label{eq:affinity}
\end{equation}
where $W$ is a learnable weight matrix, and each entry of $A$ indicates the similarity between a pair of locations in the low-level feature and the high-level feature. The features $\mathbf{v}$ and $\mathbf{x}$ of shape $(h, w, c)$ are reshaped to $(h \times w, c)$ before matrix multiplications.
We then generate the co-attended feature $\mathbf{z}_{co}$ from the affinity matrix and the original high-level feature by
$\mathbf{z}_{co} = softmax(A) \mathbf{x}$.
%

\subsection{Gated Attention Feature Aggregation}
In order to aggregate the features from the two attention modules, we concatenate the original feature $\mathbf{x}$ from the output of the encoder $E$, the self-attended feature $\mathbf{z}_{self}$ and co-attended feature $\mathbf{z}_{co}$ to generate the final saliency map using the prediction head $D$.
Due to the variations between frames, the importance of the three features may also vary in different video frames.
Therefore, instead of directly concatenating the three features, we employ a gated aggregation mechanism to dynamically combine the features.
Specifically, we use a single convolutional layer to generate a weight for each attentive feature. The operation is formulated as:
\begin{align}
f_g = \sigma (W_g \mathbf{z} + b_g),
\end{align}
where $\sigma$ is the sigmoid function with output range $[0, 1]$, and $W_g$ and $b_g$ are the weight and bias of the convolutional layer, respectively.
$f_g$ is the weight for the attentive feature, representing the portion of information that will remain in the feature. We then obtain the weighted feature by
\begin{equation}
\tilde{\mathbf{z}} = f_g \odot \mathbf{z},
\end{equation}
where $\odot$ denotes the element-wise multiplication. The gating operation is applied on $\mathbf{z}_{self}$ and $\mathbf{z}_{co}$ separately. Then the two outputs $\tilde{\mathbf{z}}_{self}$ and $\tilde{\mathbf{z}}_{co}$ are concatenated along with $\mathbf{x}$ to produce the final saliency map.


%
%
%
%
%
%


\subsection{Contrastive Feature Learning}
In order to learn better feature representations for video salient object detection, we explore the information in the whole video to improve both intra-frame discriminability and inter-frame consistency.
Similar to \cite{Khosla_nips_20}, we integrate the contrastive learning technique into our framework in a natural way for stronger supervision. 
We use multiple positive and negative samples for each anchor to improve the sampling efficiency.
Based on the intuition that the features of foreground regions in the same video should be similar, while the foreground and background are supposed to be far away in the embedding space, we adopt the contrastive learning to help the model embed features accordingly.

For an input video, the extracted features of each frame are separated into foreground and background using the ground truth masks.
Then for each foreground region $u$ as an anchor, we select the foreground regions from other frames of the same video to form positive samples $u_+$, and choose the background regions from either the same frame or other frames as negative samples $u_-$, where $u$, $u_+$ and $u_-$ are feature vectors by performing global average pooling on the features within the defined regions.
The InfoNCE contrastive loss~\cite{info_nce} is applied on the sampled features:
\begin{align}
L_{cl} = - \log \frac{\sum\limits_{u_{+} \in Q_+} e^{u^\top \cdot u_{+} / \tau}}{\sum\limits_{u_{+} \in Q_+} e^{u^\top \cdot u_{+}/ \tau} + \sum\limits_{u_{-} \in Q_-} e^{u^\top \cdot u_{-}/ \tau}},
\label{eq:cl}
\end{align}
where $Q_+$ and $Q_-$ are the sets of positive and negative samples, $\tau$ is the temperature parameter, and `$\cdot$' denotes the dot product. 

\vspace{-2mm}
{\flushleft {\bf Hard Positive and Negative Mining.}}
The selection of positive and negative samples is crucial for learning contrastive features effectively \cite{Kalantidis_nips_20}. Therefore, we employ two criteria for mining meaningful positive and negative samples: 1) Choose foreground regions with low responses in the output of our model as hard positives to suppress false negatives. 2) Select background regions with high responses in the output map as hard negatives to deal with false positives.
The hard samples are selected based on the Mean Absolute Error (MAE) between the prediction and the ground truth mask within the foreground or background region.
With the meaningful samples chosen from the whole video, the contrastive loss applied within the same frame can help the model distinguish foreground from background, and the loss enforced across different frames will improve the temporal correspondence.




\subsection{Model Training and Implementation Details}
\vspace{-2mm}
{\flushleft {\bf Overall Objective.}}
The overall objective of the proposed method is composed of the binary cross-entropy loss and the contrastive loss defined in \eqref{eq:cl}:
\begin{align}
L = L_{bce} + L_{cl},
\label{eq:loss_all}
\end{align}
where $L_{bce}$ denotes the binary cross-entropy loss computed on the output saliency map and the ground truth mask.

In the training stage, the positives and negatives for the contrastive loss are sampled from the same video as the anchor in the minibatch.
The contrastive loss is computed on multiple frames for aggregating global information in videos to learn contrastive features in both spatial and temporal dimensions.
During testing, only the current frame is taken as input to the model, which is more efficient than the temporal modeling techniques in previous methods.

\begin{table*}[!t]
\centering
\caption{\textbf{Comparisons with the state-of-the-art video salient object detection methods.} The methods in the top group are image-based approaches, while those in the bottom group are video-based mechanisms.}
\scriptsize
\begin{tabular}{cccccccccccccc}
\toprule
\multirow{2}[2]{*}{Method}& \multirow{2}[2]{*}{Year} & \multicolumn{3}{c}{DAVIS} & \multicolumn{3}{c}{FBMS} & \multicolumn{3}{c}{ViSal} & \multicolumn{3}{c}{DAVSOD} \\
\cmidrule(lr){3-5} \cmidrule(lr){6-8} \cmidrule(lr){9-11} \cmidrule(lr){12-14}
 & & maxF $\uparrow$ & S $\uparrow$ & MAE $\downarrow$ & maxF $\uparrow$ & S $\uparrow$ & MAE $\downarrow$ & maxF $\uparrow$ & S $\uparrow$ & MAE $\downarrow$ & maxF $\uparrow$ & S $\uparrow$ & MAE $\downarrow$ \\ \midrule

C2SNet~\cite{Li_ECCV_2018_2}   & ECCV'18 & 77.1 & 81.3 & 5.2 & 78.2 & 81.1 & 7.3 & 92.4 & 92.2 & 2.3 & - & - & - \\
RAS~\cite{Chen_ECCV_2018}      & ECCV'18 & 72.9 & 78.5 & 5.7 & 80.7 & 81.6 & 7.8 & 92.5 & 93.0 & 1.9 & - & - & - \\
DGRL~\cite{Wang_CVPR_2018_3}   & ECCV'18 & 76.3 & 81.2 & 5.6 & 80.2 & 82.9 & 5.7 & 91.7 & 91.6 & 2.2 & - & - & - \\
PiCANet~\cite{Liu_CVPR_2018}   & CVPR'18 & 80.1 & 84.2 & 4.4 & 81.9 & 84.5 & 5.9 & 93.2 & 93.7 & 2.2 & - & - & - \\
F$^3$Net~\cite{Wei_AAAI_2020}  & AAAI'20 & 81.9 & 85.0 & 4.1 & 81.9 & 85.3 & 6.8 & 90.7 & 87.4 & 4.5 & 56.4 & 68.9 & 11.7 \\
MINet~\cite{Pang_CVPR_2020}    & CVPR'20 & 83.5 & 86.1 & 3.9 & 81.7 & 84.9 & 6.7 & 91.1 & 90.3 & 4.1 & 58.2 & 70.4 & 10.3 \\
GateNet~\cite{Zhao_ECCV_2020}  & ECCV'20 & 84.6 & 86.9 & 3.6 & 83.2 & 85.7 & 6.5 & 92.8 & 92.1 & 3.9 & 57.8 & 70.1 & 10.4 \\
 \midrule
FCNS~\cite{Wang_TIP_2018}      & TIP'18  & 70.8 & 79.4 & 6.1 & 75.9 & 79.4 & 9.1 & 85.2 & 88.1 & 4.8 & - & - & - \\
FGRNE~\cite{Li_CVPR_2018_3}    & CVPR'18 & 78.9 & 83.6 & 3.5 & 77.8 & 80.7 & 6.9 & 84.6 & 85.8 & 4.6 & 57.3 & 69.3 & 9.8 \\
MBN~\cite{Li_ECCV_2018}        & ECCV'18 & 86.1 & 88.7 & 3.1 & 81.6 & 85.7 & 4.7 & 88.3 & 89.8 & 2.0 & - & - & - \\
PDB~\cite{Song_ECCV_2018}      & ECCV'18 & 85.2 & 88.3 & 2.9 & 81.7 & 85.2 & 6.8 & 91.1 & 90.5 & 2.9 & 57.2 & 69.8 & 11.6 \\
SSAV~\cite{Fan_CVPR_2019}      & CVPR'19 & 86.2 & 89.5 & 2.7 & 86.6 & 88.2 & 4.1 & 94.0 & 94.4 & 1.8 & 60.3 & 72.4 & 9.2 \\
AnDiff~\cite{Yang_ICCV_2019_3} & ICCV'19 & 80.8 & -    & 4.4 & 81.2 & -    & 6.4 & 90.4 & -    & 3.0 & - & - & - \\
MGA~\cite{Li_ICCV_2019}        & ICCV'19 & 90.1 & 91.1 & 2.4 & 89.8 & 90.4 & 2.9 & 94.5 & 94.3 & 1.6 & 61.8 & 72.6 & 9.0 \\
PCSA~\cite{Gu_AAAI_2020}       & AAAI'20 & 88.1 & 90.4 & 2.3 & 83.3 & 86.7 & 4.2 & 94.1 & 94.4 & 1.8 & 65.5 & 74.1 & 8.6 \\
DFNet~\cite{Zhen_ECCV_2020}    & ECCV'20 & 89.9 & -    & 1.8 & 83.3 & -    & 5.4 & 92.7 & -    & 1.7 & - & - & - \\
Ours & & \textbf{90.9} & \textbf{91.8} & \textbf{1.5} & \textbf{91.5} & \textbf{90.9} & \textbf{2.6} & \textbf{95.1} & \textbf{94.7} & \textbf{1.3} & \textbf{66.2} & \textbf{75.3} & \textbf{8.3} \\ \bottomrule
\label{tab:saliency}
\end{tabular}
\vspace{-6mm}
\end{table*}

\vspace{-2mm}
{\flushleft {\bf Implementation Details.}}
In our framework, we adopt the fully convolutional DeepLabv3~\cite{deeplabv3} as the feature encoder, which consists of the first five convolution blocks from ResNet-101~\cite{He_CVPR_2016} and an atrous spatial pyramid pooling module.
In the original implementation of DeepLabv3, the output saliency map is bilinearly upsampled by a factor of 8 to the size of the input image, which may lose detailed information.
In order to capture detailed boundaries of objects, instead of directly upsampling the saliency map to the target size, we gradually recover the spatial information with skip connections from low-level features in a way similar to \cite{Jin_NIPS_2017}.
Specifically, we upsample the map three times, each by a factor of 2. The low-level feature map of the same spatial size as the current map is fed into a convolutional layer to produce a pixel-wise prediction. The generated prediction is then added to the current map.

The low-level feature $\mathbf{v}$ used in the cross-level co-attention module is from the final convolutional layer of the first block in ResNet-101, and the high-level feature $\mathbf{x}$ used in both attention modules is from the final convolutional layer of the fourth block.
In the contrastive learning scheme, $u$, $u_+$ and $u_-$ are features before the final output layer.
In the contrastive loss (\ref{eq:cl}), the temperature $\tau$ is set to 0.1. For each anchor, we use 4 positive and 4 negative samples for optimization.
%
%
We implement the proposed model in PyTorch with the SGD optimizer. The batch size is set to 8, and the learning rate is set to $10^{-4}$.
The source code and trained models will be made available to the public.


\vspace{-1mm}
\section{Experimental Results}
\vspace{-1mm}
We evaluate the proposed framework on numerous datasets for the video salient object detection and unsupervised video object segmentation tasks. We first compare the performance of the proposed algorithm with state-of-the-art methods, and then present the ablation study to show the effectiveness of each component.
More results and analysis are provided in the supplementary material.

\subsection{Datasets and Evaluation Metrics}
\vspace{-2mm}
{\flushleft {\bf Datasets.}}
We initialize the ResNet-101 backbone with weights pre-trained on ImageNet. Following \cite{Song_ECCV_2018,Lu_CVPR_2019,Wang_ICCV_2019,Zhen_ECCV_2020}, we train the entire model on image data from MSRA10K~\cite{Cheng_PAMI_2015}, DUT-OMRON~\cite{Yang_CVPR_2013}, and video data from the training set of DAVIS~\cite{DAVIS2016}.
For video salient object detection, we evaluate the performance on the validation set of DAVIS, FBMS~\cite{Brox_ECCV_2010}, ViSal~\cite{Wang_TIP_2015} and DAVSOD~\cite{Fan_CVPR_2019} datasets.
We also report the results for unsupervised video object segmentation on the DAVIS dataset.
%

We adopt the \textbf{DAVIS} 2016 dataset, which contains 30 videos for training and 20 videos for validation with 
pixel-wise annotations in each frame.
The \textbf{FBMS} dataset is another benchmark for video object segmentation. It is composed of 29 training videos and 30 testing videos, where the video sequences are sparsely labeled.
The \textbf{ViSal} dataset is a video salient object detection dataset that consists of 17 video sequences. There are 193 frames manually annotated in the dataset.
The \textbf{DAVSOD} dataset is the current largest dataset for video salient object detection. It contains 226 videos with 23,938 frames. The pixel-wise per-frame annotations are generated according to eye fixation records.

\vspace{-2mm}
{\flushleft {\bf Evaluation Metrics.}}
To evaluate the performance of video salient object detection, we adopt the metrics F-measure, S-measure~\cite{Fan_ICCV_2017} and Mean Absolute Error (MAE).
%
%
The \textbf{F-measure} is computed upon the precision and recall:
%
$F_\beta = \frac{(1 + \beta^2) \times precision \times recall}{\beta^2 \times precision + recall}$,
%
where $\beta^2$ is set to 0.3 as suggested in \cite{Achanta_CVPR_2009_2}.
The \textbf{S-measure} evaluates the object-aware $S_o$ and region-aware $S_r$ structural similarity between the saliency map and the ground truth mask:
%
$S = \alpha S_o + (1 - \alpha) S_r$,
%
where $\alpha$ is set to 0.5.
The \textbf{MAE} measures the average of pixel-wise difference between the saliency map and the ground truth mask:
%
$MAE = \frac{\sum_{i = 1}^{H} \sum_{j = 1}^{W} | P(i, j) - G(i, j) |}{H \times W}$,
%
where $H$ and $W$ are the height and width of the image.

For the unsupervised video object segmentation task, we adopt the official evaluation metrics from the DAVIS dataset, including the region similarity $\mathcal{J}$, which is the intersection-over-union between the prediction and ground truth, and the boundary accuracy $\mathcal{F}$, which is the F-measure defined on the boundary points.


%
\begin{figure*}[t]
	\centering
	\begin{tabular}
		{ @{\hspace{0mm}}c@{\hspace{1mm}} @{\hspace{0mm}}c@{\hspace{1mm}} @{\hspace{0mm}}c@{\hspace{1mm}} @{\hspace{0mm}}c@{\hspace{1mm}} @{\hspace{0mm}}c@{\hspace{1mm}} @{\hspace{0mm}}c@{\hspace{0mm}} }
		
		\includegraphics[width=0.15\linewidth]{figure/DAVIS/breakdance_00005.jpg} &
		\includegraphics[width=0.15\linewidth]{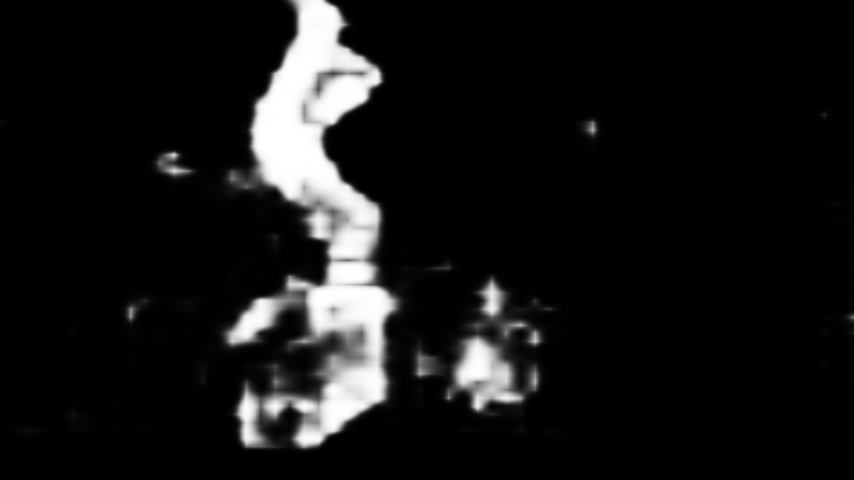} &
		\includegraphics[width=0.15\linewidth]{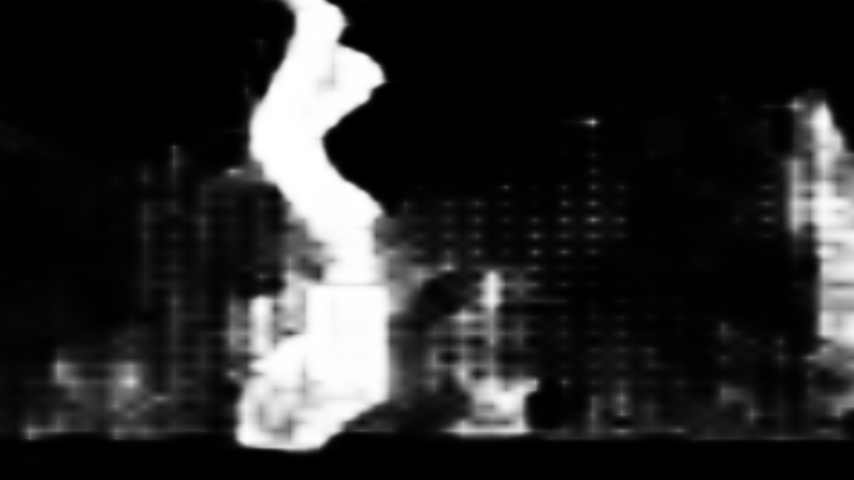} &
		\includegraphics[width=0.15\linewidth]{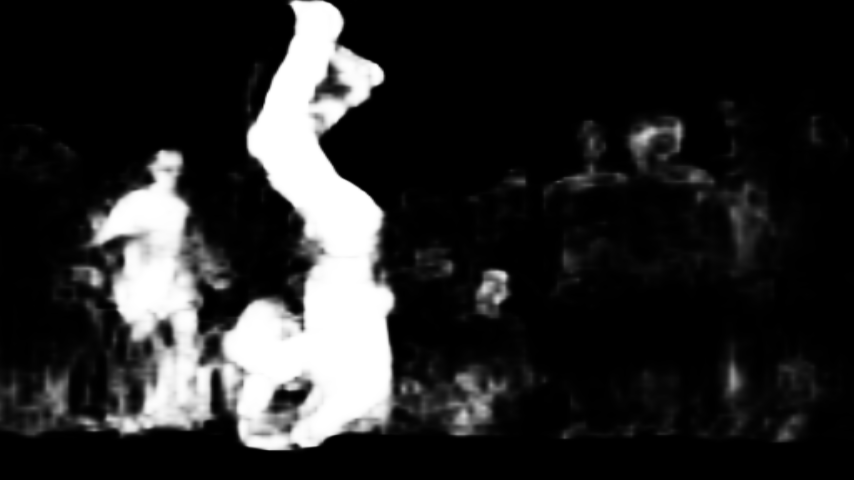} &
		\includegraphics[width=0.15\linewidth]{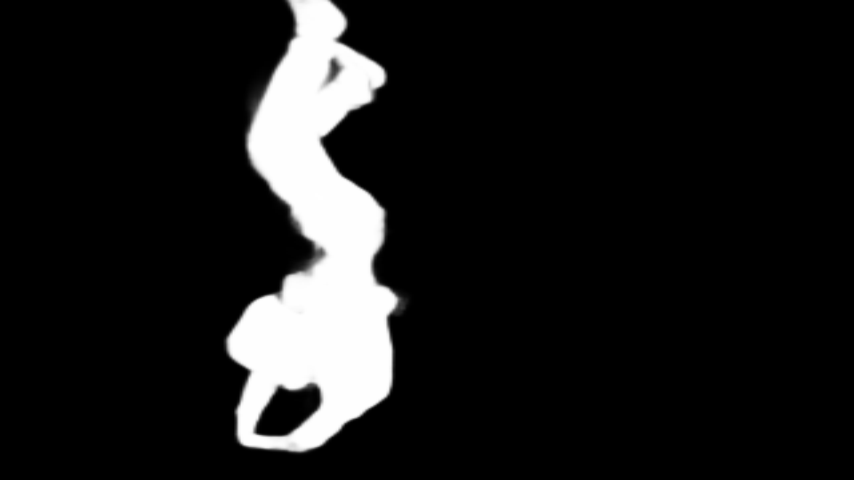} &
		\includegraphics[width=0.15\linewidth]{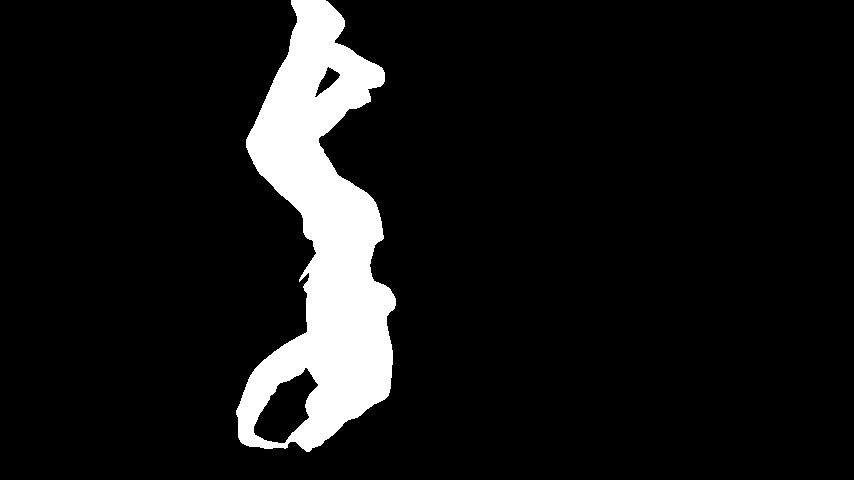} \\
		
		\includegraphics[width=0.15\linewidth]{figure/DAVIS/camel_00065.jpg} &
		\includegraphics[width=0.15\linewidth]{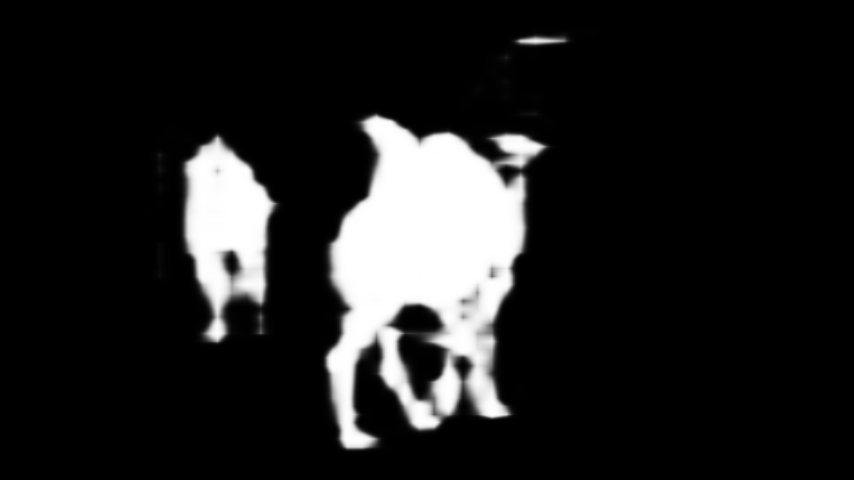} &
		\includegraphics[width=0.15\linewidth]{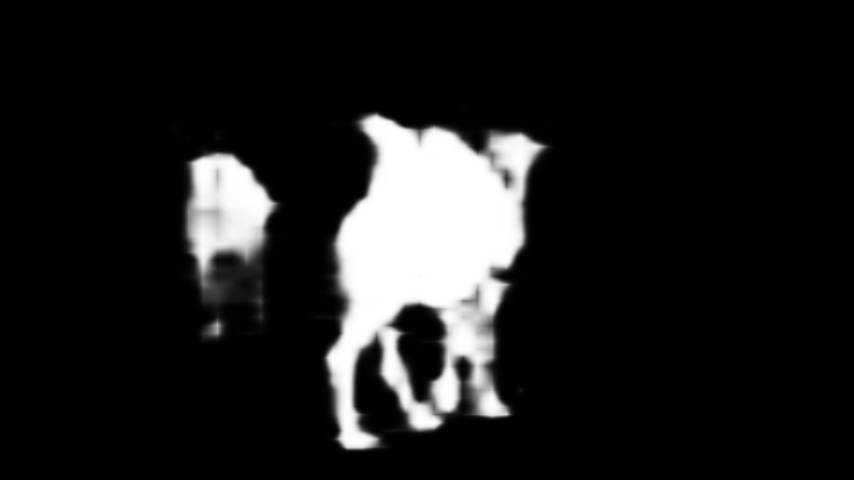} &
		\includegraphics[width=0.15\linewidth]{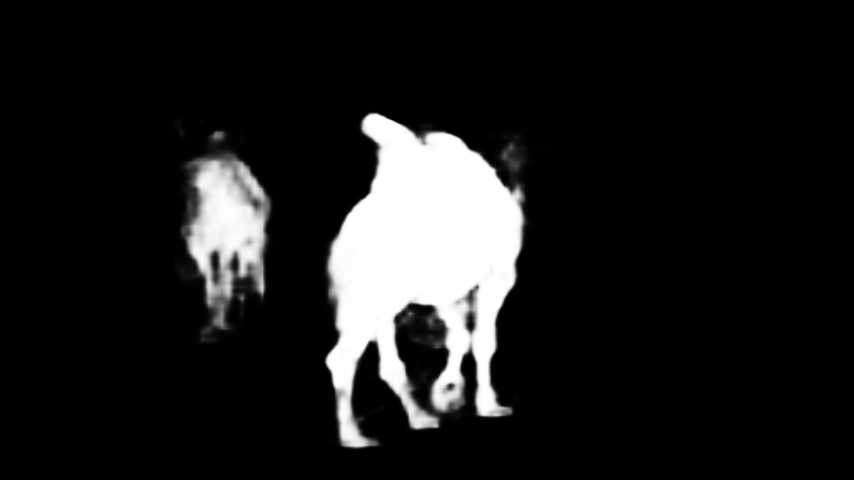} &
		\includegraphics[width=0.15\linewidth]{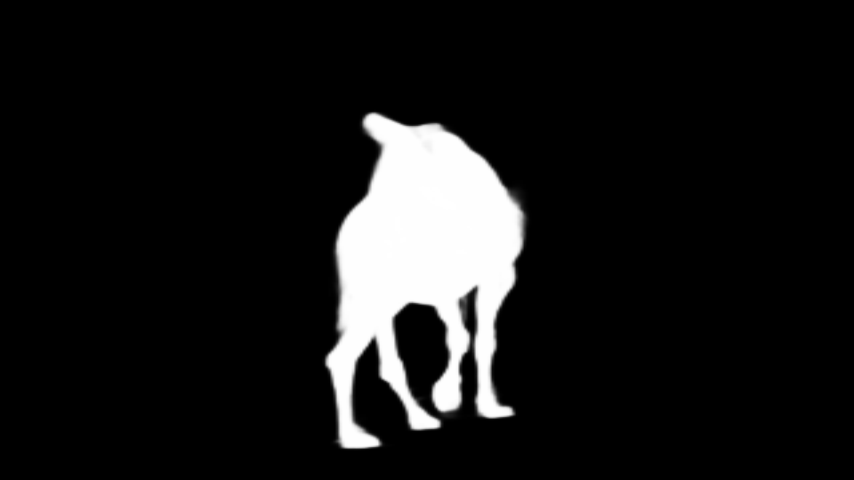} &
		\includegraphics[width=0.15\linewidth]{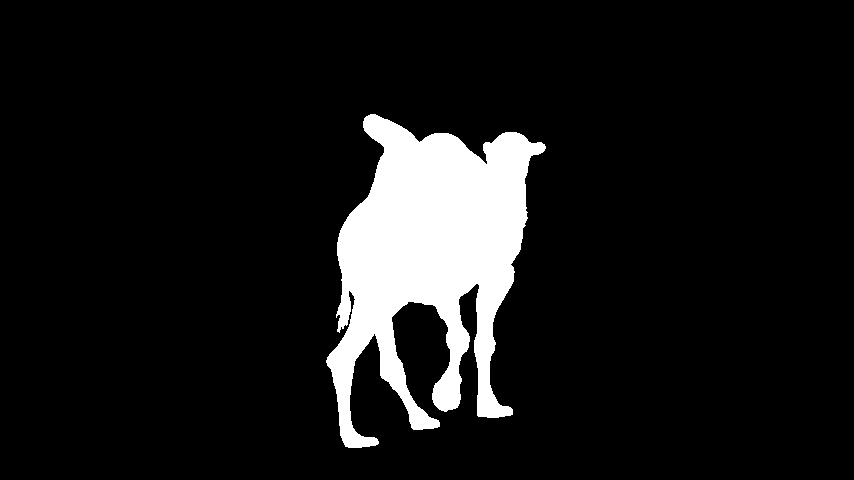} \\
		
		\includegraphics[width=0.15\linewidth]{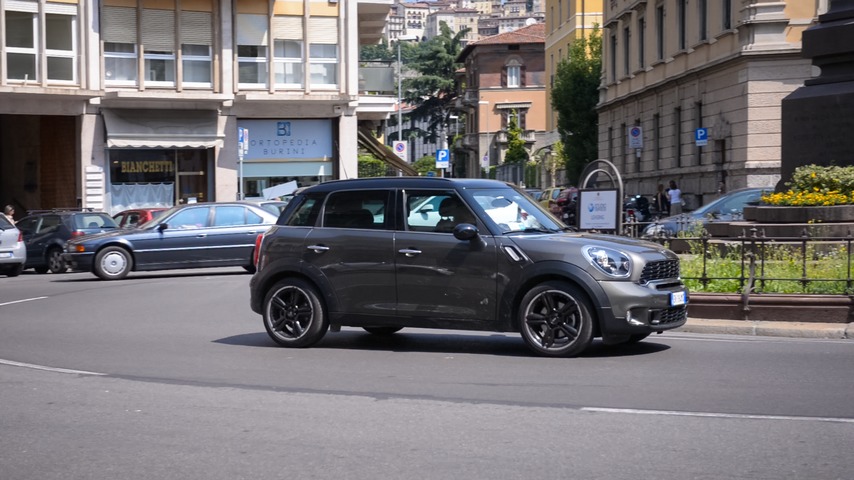} &
		\includegraphics[width=0.15\linewidth]{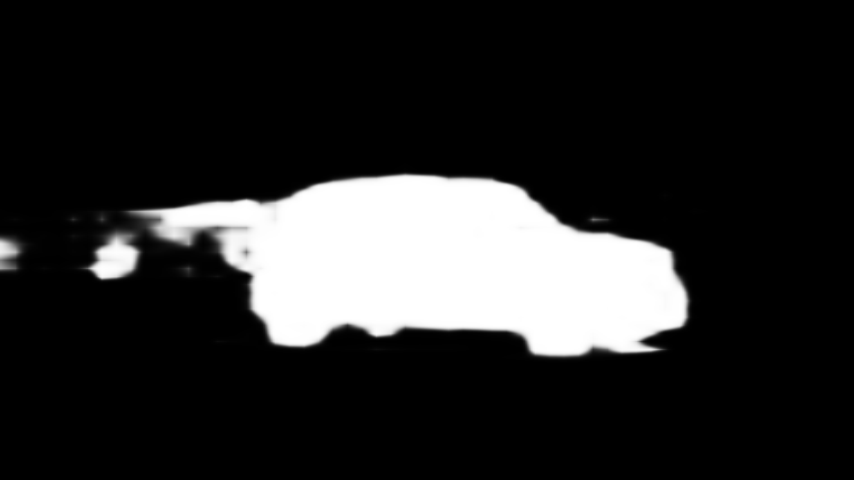} &
		\includegraphics[width=0.15\linewidth]{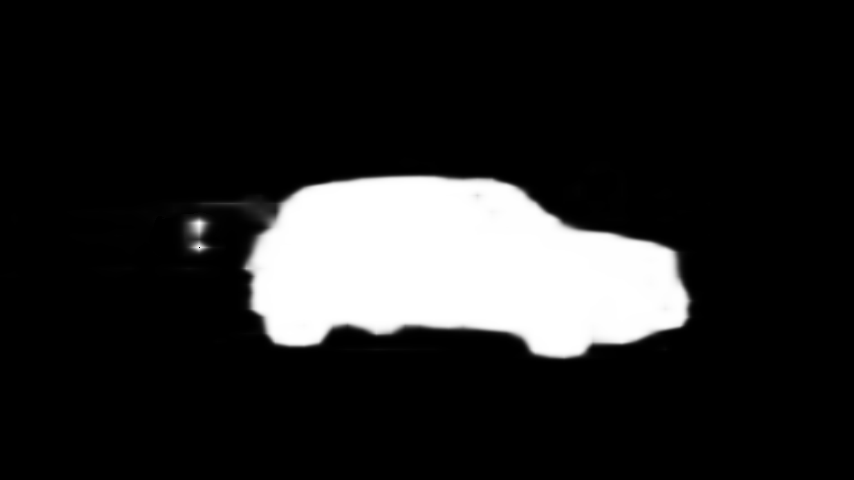} &
		\includegraphics[width=0.15\linewidth]{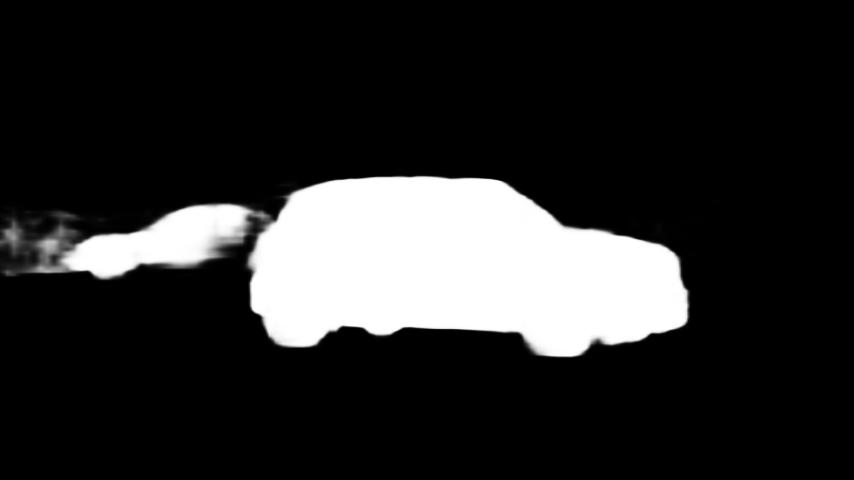} &
		\includegraphics[width=0.15\linewidth]{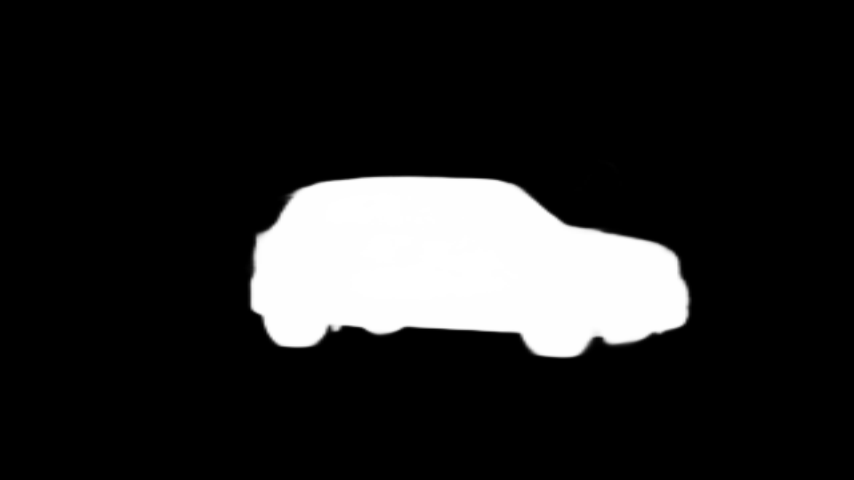} &
		\includegraphics[width=0.15\linewidth]{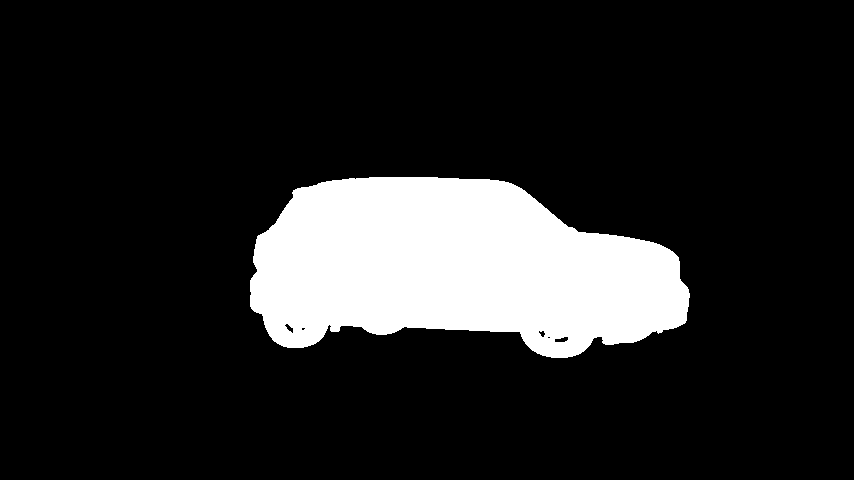} \\
		
		\includegraphics[width=0.15\linewidth]{figure/DAVIS/dance-twirl_00000.jpg} &
		\includegraphics[width=0.15\linewidth]{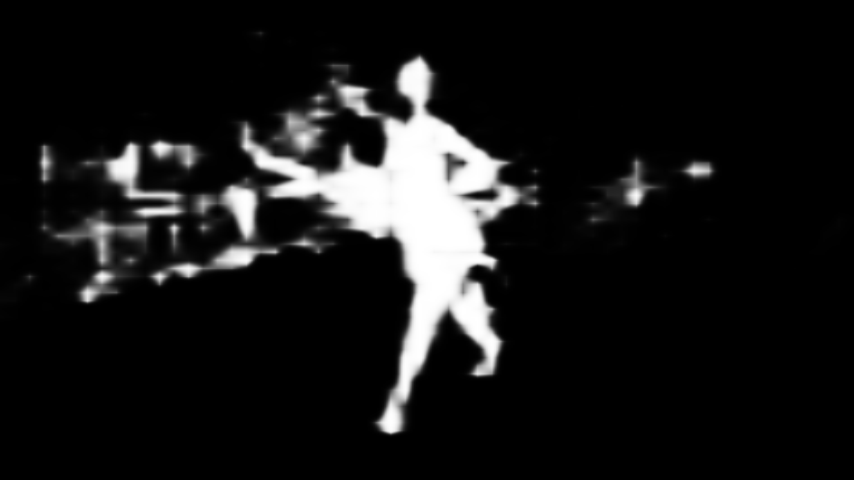} &
		\includegraphics[width=0.15\linewidth]{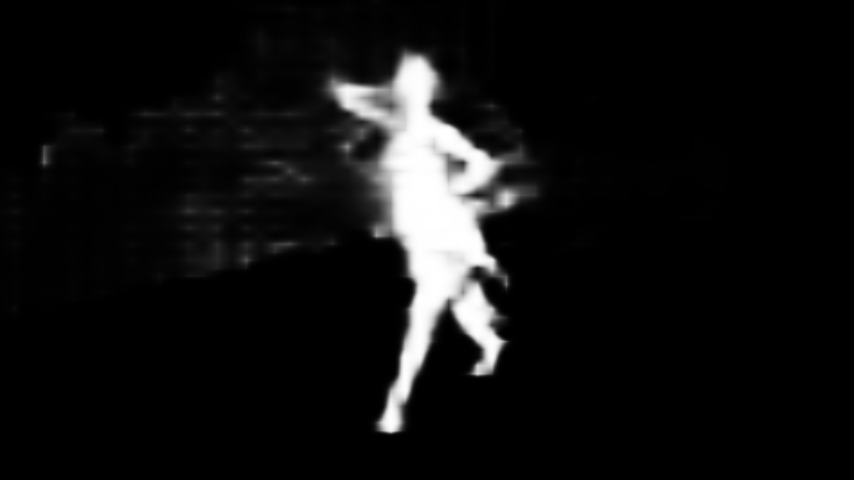} &
		\includegraphics[width=0.15\linewidth]{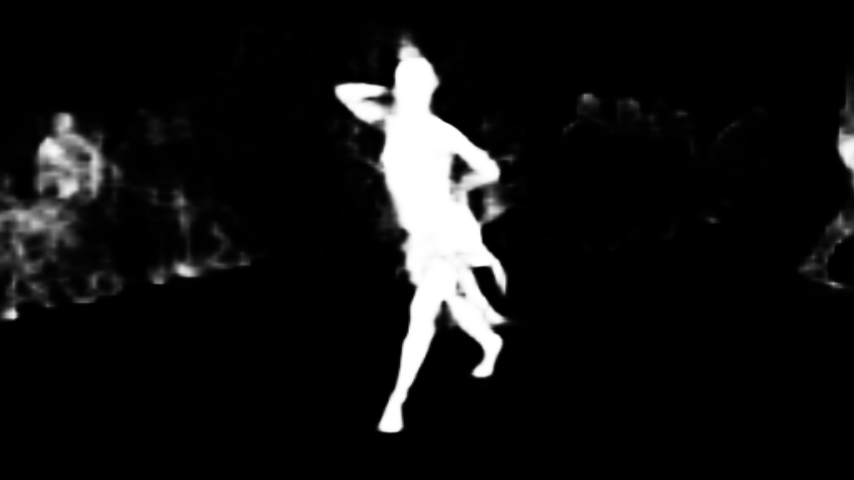} &
		\includegraphics[width=0.15\linewidth]{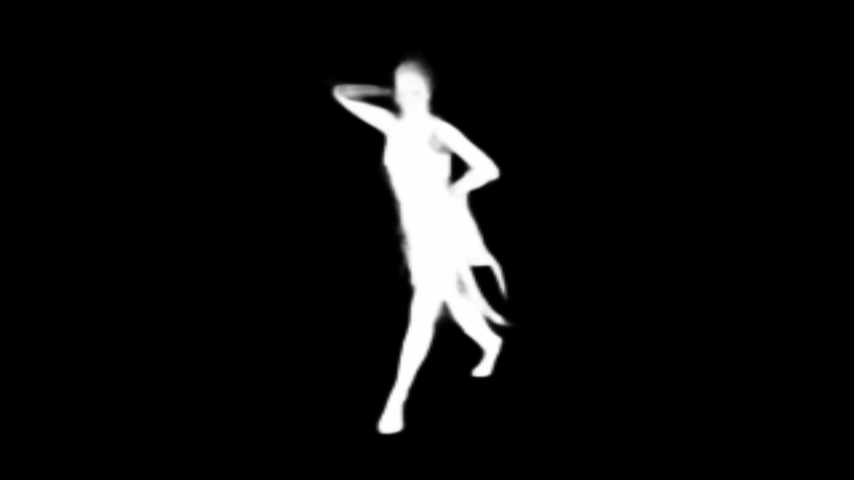} &
		\includegraphics[width=0.15\linewidth]{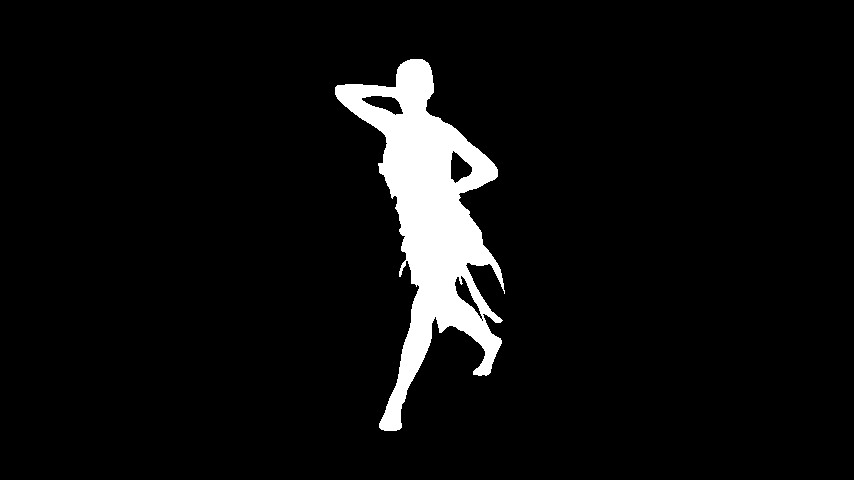} \\
		
		\includegraphics[width=0.15\linewidth]{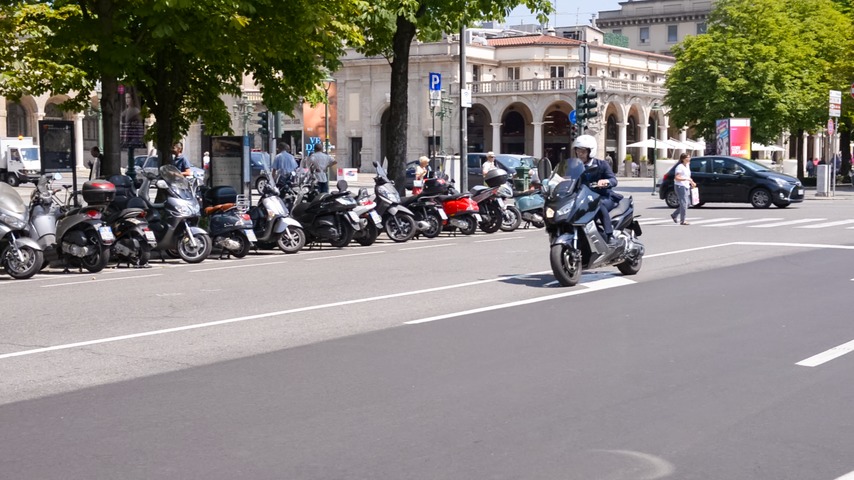} &
		\includegraphics[width=0.15\linewidth]{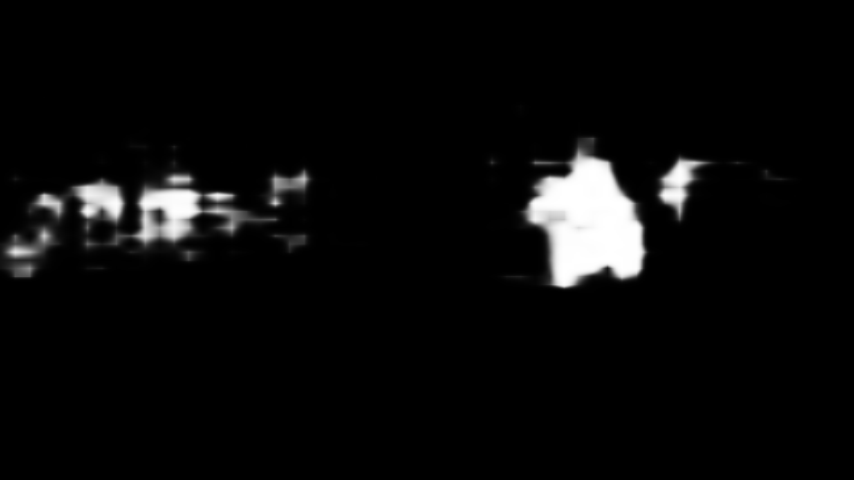} &
		\includegraphics[width=0.15\linewidth]{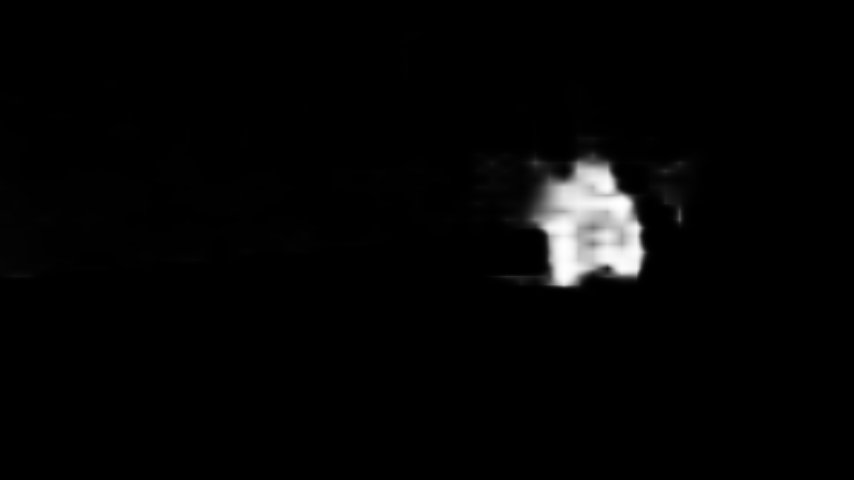} &
		\includegraphics[width=0.15\linewidth]{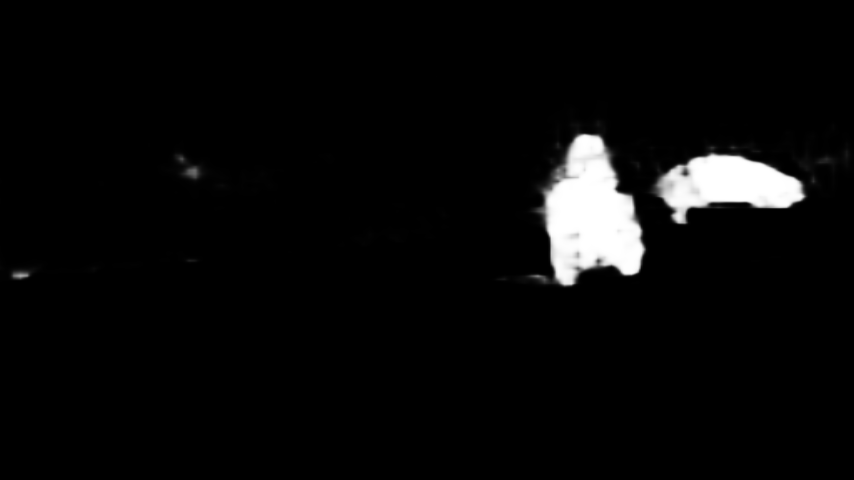} &
		\includegraphics[width=0.15\linewidth]{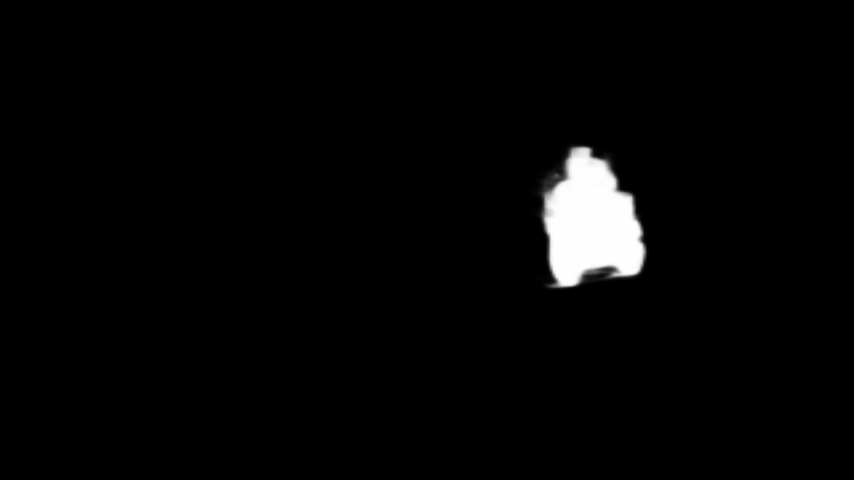} &
		\includegraphics[width=0.15\linewidth]{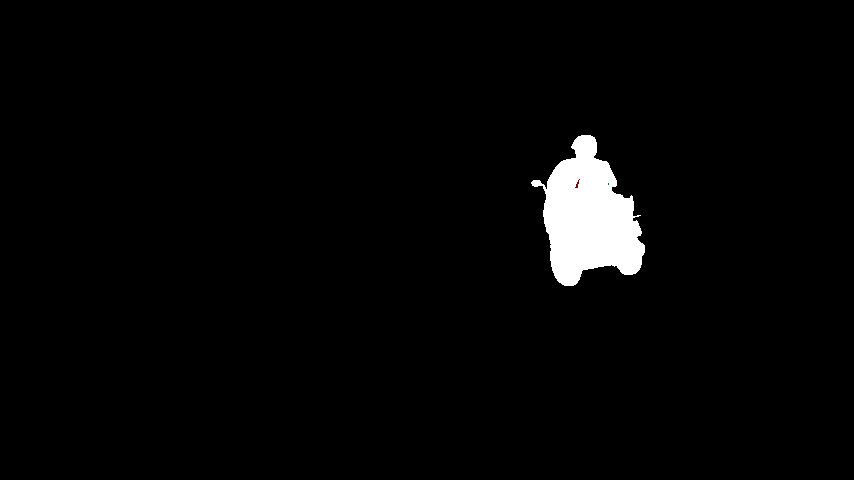} \\
		
		Input & PDB~\cite{Song_ECCV_2018} & SSAV~\cite{Fan_CVPR_2019} & MGA~\cite{Li_ICCV_2019} & Ours & GT \\
		
	\end{tabular}
	\vspace{1mm}
	\caption{\textbf{Visual comparisons with the state-of-the-art video salient object detection methods on the DAVIS dataset.} The ground truth masks (GT) are shown in the last column. The results by our method are more accurate and contain more details.}
	\label{fig:davis}
	\vspace{-4mm}
\end{figure*}

\subsection{Video Salient Object Detection}
In Table~\ref{tab:saliency}, we evaluate the proposed algorithm against the state-of-the-art salient object detection methods on the DAVIS, FBMS, ViSal and DAVSOD datasets. The top group shows the image-based approaches \cite{Li_ECCV_2018_2,Chen_ECCV_2018,Wang_CVPR_2018_3,Liu_CVPR_2018,Wei_AAAI_2020,Pang_CVPR_2020,Zhao_ECCV_2020}, and the bottom group presents the video-based methods \cite{Wang_TIP_2018,Li_CVPR_2018_3,Li_ECCV_2018,Song_ECCV_2018,Fan_CVPR_2019,Yang_ICCV_2019_3,Li_ICCV_2019,Gu_AAAI_2020,Zhen_ECCV_2020}.
We observe that video-based mechanisms generally perform better than those based on images as the image-based approaches do not consider temporal correspondence.
Compared to video-based methods that model the temporal information by optical flow \cite{Li_CVPR_2018_3,Li_ECCV_2018,Li_ICCV_2019}, ConvLSTM \cite{Li_CVPR_2018_3,Song_ECCV_2018,Fan_CVPR_2019}, or computing the relationships of the current frame and one \cite{Wang_TIP_2018,Yang_ICCV_2019_3} or multiple \cite{Gu_AAAI_2020,Zhen_ECCV_2020} reference frames, our approach achieves higher performance without the time-consuming temporal modeling techniques.
Since we learn contrastive features from the whole video, the model is able to generate results with better intra-frame discriminability and inter-frame consistency in a more efficient way.
Overall, the proposed method performs favorably against previous image-based and video-based salient object detection approaches on the four benchmark datasets.

The visual comparisons against the state-of-the-art methods \cite{Song_ECCV_2018,Fan_CVPR_2019,Li_ICCV_2019} are presented in Figures~\ref{fig:davis}. We show that the proposed model generates more accurate results with more detailed information, while the results of previous approaches often contain background regions or lose the information near boundaries.

\begin{table}[tp]
\centering
\caption{\textbf{Comparisons with the state-of-the-art unsupervised video object segmentation methods.}}
\scriptsize
\begin{tabular}{cccc}
\hline
	\multirow{2}[2]{*}{Method}& \multirow{2}[2]{*}{Year}& \multicolumn{2}{c}{DAVIS} \\
\cmidrule(lr){3-4}
 &  & $\mathcal{J}$ Mean $\uparrow$ & $\mathcal{F}$ Mean $\uparrow$ \\ \midrule
MBN~\cite{Li_ECCV_2018}        & ECCV'18 & 80.4 & 78.5 \\
MotAdapt~\cite{Siam_ICRA_2019} & ICRA'19 & 77.2 & 77.4 \\
AGS~\cite{Wang_CVPR_2019}      & CVPR'19 & 79.7 & 77.4 \\
COSNet~\cite{Lu_CVPR_2019}     & CVPR'19 & 80.5 & 79.4 \\
AGNN~\cite{Wang_ICCV_2019}     & ICCV'19 & 80.7 & 79.1 \\
AnDiff~\cite{Yang_ICCV_2019_3} & ICCV'19 & 81.7 & 80.5 \\
MGA~\cite{Li_ICCV_2019}        & ICCV'19 & 81.4 & 81.0 \\
EpO+~\cite{Faisal_WACV_2020}   & WACV'20 & 80.6 & 75.5 \\
MATNet~\cite{Zhou_AAAI_2020}   & AAAI'20 & 82.4 & 80.7 \\
GateNet~\cite{Zhao_ECCV_2020}  & ECCV'20 & 80.9 & 79.4 \\
DFNet~\cite{Zhen_ECCV_2020}    & ECCV'20 & 83.4 & 81.8 \\
Ours & & \textbf{83.5} & \textbf{82.0} \\ \hline
\label{tab:segmentation}
\end{tabular}
\vspace{-6mm}
\end{table}
\begin{table*}[tp]
\centering
\caption{\textbf{Ablation study of the proposed method.} We show the effectiveness of each component in the proposed framework, including the cross-level co-attention (co-attn), non-local self-attention (self-attn), contrastive loss ($L_{cl}$), and hard sample mining.}
\setlength\tabcolsep{3pt}
\scriptsize
\begin{tabular}{lcccccccccccc}
\toprule
\multirow{2}[2]{*}{Method}& \multicolumn{3}{c}{DAVIS} & \multicolumn{3}{c}{FBMS} & \multicolumn{3}{c}{ViSal} & \multicolumn{3}{c}{DAVSOD} \\
\cmidrule(lr){2-4} \cmidrule(lr){5-7} \cmidrule(lr){8-10} \cmidrule(lr){11-13}
 & maxF $\uparrow$ & S $\uparrow$ & MAE $\downarrow$ & maxF $\uparrow$ & S $\uparrow$ & MAE $\downarrow$ & maxF $\uparrow$ & S $\uparrow$ & MAE $\downarrow$ & maxF $\uparrow$ & S $\uparrow$ & MAE $\downarrow$ \\ \midrule
Baseline                       & 86.8 & 88.4 & 3.1 & 86.9 & 84.2 & 5.8 & 92.4 & 91.8 & 2.9 & 62.9 & 72.1 & 9.6 \\
Baseline + co-attn             & 88.3 & 89.7 & 2.8 & 88.6 & 86.3 & 5.2 & 93.7 & 92.6 & 2.5 & 64.2 & 73.4 & 9.2 \\
Baseline + self-attn           & 88.5 & 90.3 & 2.7 & 89.0 & 86.6 & 5.3 & 93.9 & 93.0 & 2.3 & 64.5 & 73.6 & 9.1 \\
Baseline + self-attn + co-attn & 89.3 & 90.7 & 2.1 & 90.6 & 89.1 & 4.3 & 94.5 & 93.6 & 1.9 & 65.4 & 74.4 & 8.7 \\
Baseline + self-attn + co-attn + $L_{cl}$ & 89.9 & 91.1 & 1.8 & 91.0 & 89.8 & 3.8 & 94.8 & 94.0 & 1.7 & 65.9 & 75.0 & 8.4 \\
Baseline + self-attn + co-attn + $L_{cl}$ + hard mining & \textbf{90.9} & \textbf{91.8} & \textbf{1.5} & \textbf{91.5} & \textbf{90.9} & \textbf{2.6} & \textbf{95.1} & \textbf{94.7} & \textbf{1.3} & \textbf{66.2} & \textbf{75.3} & \textbf{8.3} \\ \bottomrule
\label{tab:ablation}
\end{tabular}
\vspace{-4mm}
\end{table*}
\begin{table*}[t]
\centering
\caption{\textbf{Comparisons of temporal modeling techniques.} We compare our method with the co-attention technique in COSNet~\cite{Lu_CVPR_2019}.}
\scriptsize
\begin{tabular}{cccccccccccccc}
\toprule
\multirow{2}[2]{*}{Method} & \multicolumn{3}{c}{DAVIS} & \multicolumn{3}{c}{FBMS} & \multicolumn{3}{c}{ViSal} & \multicolumn{3}{c}{DAVSOD} & \multirow{2}[2]{*}{Time (s)} \\
\cmidrule(lr){2-4} \cmidrule(lr){5-7} \cmidrule(lr){8-10} \cmidrule(lr){11-13}
 & maxF $\uparrow$ & S $\uparrow$ & MAE $\downarrow$ & maxF $\uparrow$ & S $\uparrow$ & MAE $\downarrow$ & maxF $\uparrow$ & S $\uparrow$ & MAE $\downarrow$ & maxF $\uparrow$ & S $\uparrow$ & MAE $\downarrow$ & \\ \midrule
Baseline       & 86.8 & 88.4 & 3.1 & 86.9 & 84.2 & 5.8 & 92.4 & 91.8 & 2.9 & 62.9 & 72.1 & 9.6 & 0.11 \\
+ co-attn~\cite{Lu_CVPR_2019} & 88.1 & 89.3 & 2.9 & 88.4 & 86.0 & 5.1 & 93.5 & 92.5 & 2.5 & 64.6 & 74.1 & 8.7 & 0.45 \\
Ours & \textbf{90.9} & \textbf{91.8} & \textbf{1.5} & \textbf{91.5} & \textbf{90.9} & \textbf{2.6} & \textbf{95.1} & \textbf{94.7} & \textbf{1.3} & \textbf{66.2} & \textbf{75.3} & \textbf{8.3} & 0.26 \\ \bottomrule
\label{tab:co_attn}
\end{tabular}
\vspace{-4mm}
\end{table*}

\begin{table*}[!t]
\centering
\caption{\textbf{Comparisons with the state-of-the-art methods on computational complexity.}}
\scriptsize
\begin{tabular}{ccccccc}
\toprule
& COSNet~\cite{Lu_CVPR_2019} & AGNN~\cite{Wang_ICCV_2019} & AnDiff~\cite{Yang_ICCV_2019_3} & MGA~\cite{Li_ICCV_2019} & DFNet~\cite{Zhen_ECCV_2020} & Ours \\ \midrule
\# Params (M)  & 81.2   & 82.3   & 79.3  & 254.0  & 64.7 & \textbf{59.3}  \\
FLOPs (G)      & 1148.5 & 1921.0 & 726.8 & 2265.7 & -    & \textbf{425.7} \\
Runtime (s)    & 0.45 & 0.55 & 0.36 & 0.29 & - & \textbf{0.26} \\ \bottomrule

\label{tab:computation}
\end{tabular}
\vspace{-6mm}
\end{table*}

\subsection{Unsupervised Video Object Segmentation}
In addition to the video salient object detection task, we evaluate the proposed method on the unsupervised video object segmentation task as a downstream application.
Table~\ref{tab:segmentation} shows the evaluation results against the state-of-the-art approaches \cite{Li_ECCV_2018,Siam_ICRA_2019,Wang_CVPR_2019,Lu_CVPR_2019,Wang_ICCV_2019,Yang_ICCV_2019_3,Li_ICCV_2019,Faisal_WACV_2020,Zhou_AAAI_2020,Zhao_ECCV_2020,Zhen_ECCV_2020} on the DAVIS dataset.
To capture temporal cues, prior methods employ optical flow \cite{Li_ECCV_2018,Siam_ICRA_2019,Li_ICCV_2019,Faisal_WACV_2020,Zhou_AAAI_2020}, ConvLSTM \cite{Wang_CVPR_2019}, graph neural networks \cite{Wang_ICCV_2019}, or calculate the correlation between the current frame and other reference frames \cite{Lu_CVPR_2019,Yang_ICCV_2019_3,Zhen_ECCV_2020}.
While these techniques for modeling temporal information usually entail high computational cost, we adopt a more efficient way by learning contrastive features across video frames.
The proposed algorithm outperforms most methods by significant margins, and is competitive with the DFNet~\cite{Zhen_ECCV_2020} scheme.
We note that DFNet uses multiple frames as input during testing, which requires additional computation. In contrast, we do not use information from other frames in the testing stage, which is much more efficient.
Furthermore, we show the merits of the proposed method over DFNet on the saliency task in Table~\ref{tab:saliency}.

\subsection{Ablation Study}
In Table~\ref{tab:ablation}, we provide the ablation study results to show the contribution of each component in the proposed method. 
We first present the results of the baseline model that does not contain the attention modules. By integrating either the cross-level co-attention or non-local self-attention module, the performance improves from the baseline.
When both attention modules are incorporated into the network, the results are further improved. The performance gains made by the two attention modules show the merits of aggregating features from different perspectives.
With the contrastive loss $L_{cl}$ added to the model, the performance is improved consistently.
Finally, the full model that utilizes hard sample mining for contrastive learning achieves the best performance, demonstrating that the contrastive learning technique along with the well-designed sampling strategy helps the model learn better feature representations for video salient object detection.

\subsection{Discussions on Temporal Modeling}
%
As we aim to avoid the temporal modeling techniques that require high computational cost, we learn the temporal dependency information during training via the contrastive loss.
While we do not explicitly model temporal information during testing, the contrastive loss helps the model learn transformation invariant features of the commonly existing objects in different background. 
Thus, the learned features are more robust to distribution shifts, and can improve the temporal consistency during testing.
%
%
For fair comparisons with other schemes that utilize temporal information in the testing stage, we incorporate into our baseline the technique in COSNet~\cite{Lu_CVPR_2019}, which performs co-attention of two randomly sampled frames during training. During the testing stage, co-attention is applied between the current frame and five other randomly sampled frames.
We train the model with the default settings as COSNet~\cite{Lu_CVPR_2019}.
%
The performance and runtime are shown in Table~\ref{tab:co_attn}. We observe that the proposed model outperforms their mechanism, while the co-attention technique in COSNet does not improve much from the baseline but largely increases the runtime due to the computation on multiple frames.

\subsection{Analysis of Computational Complexity}
%
Table~\ref{tab:computation} shows the specifications of the state-of-the-art methods \cite{Lu_CVPR_2019,Wang_ICCV_2019,Yang_ICCV_2019_3,Li_ICCV_2019,Zhen_ECCV_2020} on the number of parameters, floating point operations (FLOPs) and runtime, where the FLOPs and runtime are only computed on methods with publicly available models, using 
a machine with an NVIDIA GTX 1080 Ti GPU.
Compared to other methods, our model has the least parameters.
The AGNN~\cite{Wang_ICCV_2019} model has significantly high computational complexity due to the usage of graph neural networks.
The COSNet~\cite{Lu_CVPR_2019} and AnDiff~\cite{Yang_ICCV_2019_3} schemes take one or multiple additional frames as a reference to generate the output. 
As a result, they also require high computational cost.
Similar to COSNet~\cite{Lu_CVPR_2019}, DFNet~\cite{Zhen_ECCV_2020} also adopts multiple frames during inference to capture more information.
The MGA~\cite{Li_ICCV_2019} approach predicts the output from an RGB image and an optical flow map, where the optical flow is estimated by an external model. We include the computation of flow estimation (FlowNet 2.0~\cite{Ilg_CVPR_2017} as reported in the paper), where a large number of computations are contributed by the flow model (2019 G out of 2265.7 G).
Since our model only uses the current frame as input and does not employ optical flow during inference, the computation is much more efficient than other mechanisms. 
Furthermore, our model requires the least runtime.


\vspace{-1mm}
\section{Conclusions}
\vspace{-1mm}
In this paper, we propose an end-to-end trainable framework for video salient object detection with \emph{higher accuracy}, \emph{smaller model size} and \emph{lower runtime} than previous methods.
Our model contains two attention modules to aggregate features from different perspectives.
First, the non-local self-attention technique captures long-distance correspondence by directly computing the relationships between any two locations.
Second, the cross-level co-attention mechanism computes the correlations between different feature levels to incorporate the high-resolution details into the semantic features.
In addition, we apply the contrastive learning technique with the well-designed sampling strategy to learn features with better intra-frame discriminability and inter-frame consistency.
%
%
%
%
Extensive experiments and ablation study demonstrate the effectiveness of our algorithm.

{\small
\bibliographystyle{ieee_fullname}
\bibliography{mybib}
}

\end{document}